\documentclass[preprint,12pt,authoryear]{elsarticle}

\usepackage{amssymb}

\usepackage{amsmath}
\usepackage{float}
\usepackage{graphicx}
\usepackage{tikz}
\usepackage{hyperref}

\usepackage{xcolor}

\newcommand{\blue}[1]{\textcolor{blue}{#1}}

\newcommand{\norm}[1]{\left\lVert#1\right\rVert}

\journal{Medical Image Analysis}

\begin{document}

\begin{frontmatter}

\title{An Ordinal Regression Framework for a Deep Learning Based Severity Assessment for Chest Radiographs}

\author[rad]{Patrick Wienholt}
\author[rad,vci]{Alexander Hermans}
\author[rad]{Firas Khader}
\author[medInf,Surg]{Behrus Puladi}
\author[vci]{Bastian Leibe}
\author[rad]{Christiane Kuhl}
\author[rad]{Sven Nebelung}
\author[rad]{Daniel Truhn}

\affiliation[rad]{organization={Department of Diagnostic and Interventional Radiology, RWTH Aachen University Hospital},%
            city={Aachen},
            country={Germany}}
\affiliation[vci]{organization={Visual Computing Institute, RWTH Aachen University},%
            city={Aachen},
            country={Germany}}
\affiliation[medInf]{organization={Institute of Medical Informatics, RWTH Aachen University Hospital},%
            city={Aachen},
            country={Germany}}
\affiliation[Surg]{organization={Department of Oral and Maxillofacial Surgery, RWTH Aachen University Hospital},%
            city={Aachen},
            country={Germany}}

\begin{abstract}
This study investigates the application of ordinal regression methods for categorizing disease severity in chest radiographs.
We propose a framework that divides the ordinal regression problem into three parts: a model, a target function, and a classification function. 
Different encoding methods, including one-hot, Gaussian, progress-bar, and our soft-progress-bar, are applied using ResNet50 and ViT-B-16 deep learning models.
We show that the choice of encoding has a strong impact on performance and that the best encoding depends on the chosen weighting of Cohen's kappa and also on the model architecture used.
We make our code publicly available on \href{https://github.com/paddyOnGithub/ordinal_regression}{\blue{GitHub}}.
\end{abstract}

\begin{keyword}

ordinal regression framework \sep chest radiographs \sep deep learning

\end{keyword}

\end{frontmatter}

\section{Introduction}
Clinical medicine often requires a nuanced assessment of a patient's health status. 
This is reflected by the fact that most radiological reports are communicated via prosaic text rather than by categorical binary labels. 
For example, if a patient suffers from pneumonia, the treating physician needs to know to what extent the patient is affected. Similarly, pleural effusion can range in severity from mild to very severe. 
A mild pleural effusion usually has little direct effect on the patient, but is still of interest because it can be an indication of conditions such as heart failure or pneumonia. 
A severe pleural effusion, on the other hand, can be a sign of cancer. 
It can also cause chest pain or dyspnea on exertion \citep{corcoran2015pleural}. 

To date, this nuanced assessment is not reflected in most data that is used to train deep learning algorithms. 
Instead, most datasets provide binary labels only. 
Consequently, most deep learning models do not provide an accurate and systematic ordinal severity scale. 

In the context of applying machine learning techniques to interpret chest radiographs, the typical approach involves binary classification, distinguishing only between the mere presence or absence of the disease.
In doing so, many researchers neglect the ordinality of the data \citep{agresti2010analysis}.
In the realm of classifying chest radiographs, widely used public datasets such as the CheXpert dataset \citep{Irvin2019Chexpert}, the VinDr-CXR dataset \citep{nguyen2022vindr} and the MIMIC-CXR-JPG dataset \citep{johnson2019mimicCxrJpg} are available. 
Notably, these datasets lack nuanced severity classifications and provide labels solely indicating whether a disease is present or absent.
On the other hand, the MIMIC-CXR \citep{Johnson2019MIMICCXR} dataset does not provide concrete labels, but only free text diagnoses that reflect different nuances, but this also makes it difficult to train models to diagnose diseases.
Therefore, we use our own dataset of 193k images which has been labeled by experienced radiologists on a graded scale. 
I.e., for every disease, the severity is given on a scale from 1-5 with 5 signifying the maximum extent of the disease and 1 denoting a minor onset of the disease.

The primary objective of this study is to comprehensively investigate the approaches of ordinal regression for effectively categorizing the disease severity in chest radiographs and to identify the most suitable approach. 
To fulfill this objective, a framework is proposed to perform ordinal regression. 
This framework entails three fundamental components: a DL model, a target function utilized for model training, and a classification function to map the output of the trained model onto the ordinal classes.

Our contributions are: 
(1.) We consolidate various methods from existing literature into a coherent framework that can also accommodate future ordinal regression approaches.
(2.) We introduce a novel ordinal regression approach termed the ``soft-progress-bar'', designed with a distinct focus on mitigating large deviations.
(3.) We conduct a systematic evaluation of different regression methods on chest radiographs, considering diverse deviation weightings of outliers to assess their performance using differently weighted Cohen's kappas.

\subsection{Related Work}
Ordinal regression is a method that has been used outside of medicine. 
An overview over methods and models that use non-neural networks was compiled by \citet{agresti2010analysis}.
Notably, they found that even when ordinal data is available, many researchers do not use this information, but treat the data as if they were nominal.
With the advent of neural networks, ordinal regression methods were also applied within this field and further refined.
\citet{gutierrez2016OrdinalRegression} provide an overview of ordinal regression methods in machine learning problems.
A taxonomy for different methods is introduced. 
In particular, for medical images, ordinal sorting of images into different severity levels is a common way to determine severity on a scale that is more diverse than simply indicating the presence or absence of a disease.
An example of this is prostate cancer using the Gleason score. 
For example, \citet{DeVente2021} perform ordinal regression to predict the Gleason score of prostate cancer.
They use the 2D U-net architecture \citep{Ronneberger2015UNet} to perform this classification based on MRI images.
Similarly, \citet{cao2019mpMRIProstata} use the Gleason score to train their FocalNet, a conventional neural network (CNN) architecture, on multiparametric MRI images.
Since one-hot encoding penalizes all deviations equally, the authors opted for an encoding a reminiscent of a progress-bar.
\citet{Horng2021} present an application of ordinal regression to detect and quantify the severity of pulmonary edema from chest radiographs.
To assess the severity of diabetic retinopathy from images, \citet{Tian2021} and \citet{Liu2018} use ordinal regression.
Ordinal regression is also used to estimate the age of individuals from MRI images \citep{He2022MRI,Feng2020}.
\citet{Niu2016} use images of faces to estimate the age of individuals on an ordinal scale. The authors encode the ordinal classes into a vector that takes the value 1 in more places depending on the ranking of the class, like a progress-bar.
A very similar encoding is also proposed by \citet{Cheng2008Ord}.
A more general approach is suggested by \citet{Berg2020LabelDiversity}.
nstead of running one regression via classification, the authors run several of them simultaneously with different thresholds, which gives them a lower error.
Another approach is proposed by \citet{Diaz2019}. 
Their proposed soft ordinal vectors correspond to a softer one-hot encoding that considers the ordinal nature of the label. 
Similarly, \citet{Tan2016} use a Gaussian function to obtain a softer encoding that considers the ordinal nature of the data.
Although there has been research on different ordinal regression approaches, there has been no direct comparison of different methods applied to chest radiographs.
Our research aims to fill this gap.

\subsection{Dataset}
The dataset we use originates from the internal repository of the Radiology Department of the University Hospital Aachen. 
It encompasses a total of 193,361 bedside chest radiographs, captured during the time span from 2009 to 2020. 
The radiographs were acquired from 54,176 patients, of which 33,521 were male, 20,653 were female, and two were labeled neither male, nor female. 
These bedside radiographs are taken from patients in the intensive care unit. 
In the context of our work, the diagnosis comprises the following conditions: pulmonary congestion, pleural effusion, pneumonic infiltrates and atelectasis. 
With the exception of pulmonary congestion, each of these conditions is associated with distinct labels designated for the left and right patient side, respectively. 
There are five discrete levels of severity for each of these seven labels. 
The severity labels were assigned directly to each of these images by radiologists. 
For the sake of clarity, the ordinal rank $k$ associated with a class will be denoted as the class in the following. 
In ascending order of severity, the $K = 5$ ordinal levels are as follows:
\begin{table}[h!]
    \begin{tabular}[h]{c|c|l}
        Label & $k$ & Description \\
        \hline
        \textit{None} & 1 & the absence of the specified condition \\
        \textit{(+)} & 2 & a vague and uncertain presence of the condition \\
        \textit{+} & 3 & a clear but relatively mild presence of the disease \\
        \textit{++} & 4 & a definite presence with a moderate degree of severity \\
        \textit{+++} & 5 & a strikingly strong manifestation of the disease \\
    \end{tabular}
    \caption[ordinal label]{
        \label{tab:ordLabel}The five different ordinal labels and their descriptions. The ordinal rank is also given by the variable $k$.
}
\end{table}

It is important to note that the assignment of the different severity classifications was at the discretion of the radiologists and thus subject to inter-reader variability.
Since there are different ways to apply ordinal regression to these images, we introduce a framework that summarizes these different approaches.

\section{The Framework}
\begin{figure}
    \centering
    \resizebox{0.96\textwidth}{!}{
    \begin{tikzpicture}[
        txt_box_green/.style={rectangle, fill=lime!50!lightgray,rounded corners=10,minimum height = 2cm,minimum width=4cm,align=center},
        txt_box_green_small/.style={rectangle, fill=lime!50!lightgray,rounded corners=10,minimum height = 1cm,minimum width=2.3cm,align=center},
        txt_box_blue/.style={rectangle, fill=cyan!50!lightgray,rounded corners=10,minimum height = 2cm,minimum width=4cm,align=center},
        txt_box_blue_small/.style={rectangle, fill=cyan!50!lightgray,rounded corners=10,minimum height = 1cm,minimum width=2.3cm,align=center},
        txt_box_model/.style={rectangle, fill=white!50!lightgray,rounded corners=10,minimum height = 2cm,minimum width=4cm,align=center},
        black_line/.style={draw=black!100,line width=1.0mm},
        bounding_box_green/.style={rectangle, draw=lime!50!lightgray,rounded corners=10,line width=1.0mm},
        bounding_box_blue/.style={rectangle, draw=cyan!50!lightgray,rounded corners=10,line width=1.0mm},
        ]
        \node[txt_box_model,text width=3.6cm] (model) at (0,0) {\textbf{Model}};
        \node[txt_box_green,text width=3.6cm] (target) at (6,1.5) {\textbf{Target Function}};
        \node[txt_box_blue,text width=3.6cm] (class) at (6,-1.5) {\textbf{Classification Function}};
        \node[txt_box_blue_small,align=center] at (9.7,-1.5) {Inference};
        \node[txt_box_green_small,align=center] at (9.7,1.5) {Training};

        \draw[bounding_box_green, dotted] ([shift={(0mm,4mm)}]model.north west) -- ([shift={(0mm,4mm)}]model.north east) -- ([shift={(0,4mm)}]target.north west) -- ([shift={(0,4mm)}]target.north east) -- ([shift={(4mm,0)}]target.north east) -- ([shift={(4mm,0)}]target.south east) -- ([shift={(0,-4mm)}]target.south east) -- ([shift={(0,-4mm)}]target.south west) -- ([shift={(0mm,-4mm)}]model.south east) -- ([shift={(0mm,-4mm)}]model.south west) -- ([shift={(-4mm,0mm)}]model.south west) -- ([shift={(-4mm,0mm)}]model.north west) -- cycle;

        \draw[bounding_box_blue, dotted] ([shift={(0mm,4mm)}]model.north west) -- ([shift={(0mm,4mm)}]model.north east) -- ([shift={(0,4mm)}]class.north west) -- ([shift={(0,4mm)}]class.north east) -- ([shift={(4mm,0)}]class.north east) -- ([shift={(4mm,0)}]class.south east) -- ([shift={(0,-4mm)}]class.south east) -- ([shift={(0,-4mm)}]class.south west) -- ([shift={(0mm,-4mm)}]model.south east) -- ([shift={(0mm,-4mm)}]model.south west) -- ([shift={(-4mm,0mm)}]model.south west) -- ([shift={(-4mm,0mm)}]model.north west) -- cycle;

        \draw [-,black_line] (model.east) to[out=0, in=180] (target.west);
        \draw [-,black_line] (model.east) to[out=0, in=180] (class.west);

    \end{tikzpicture}
}
    \caption{\label{fig:RegFramework}In our framework, the regression task is divided into three parts:
    the model trained on the task, the target function that defines a vector for each class the model is trained on, and the classification function that maps the output of the trained model to one of the possible classes.
    During training, only the model and the target function are used, while during inference only the model and the classification function are used.
    The modular framework makes it possible to exchange the three parts.}
\end{figure}
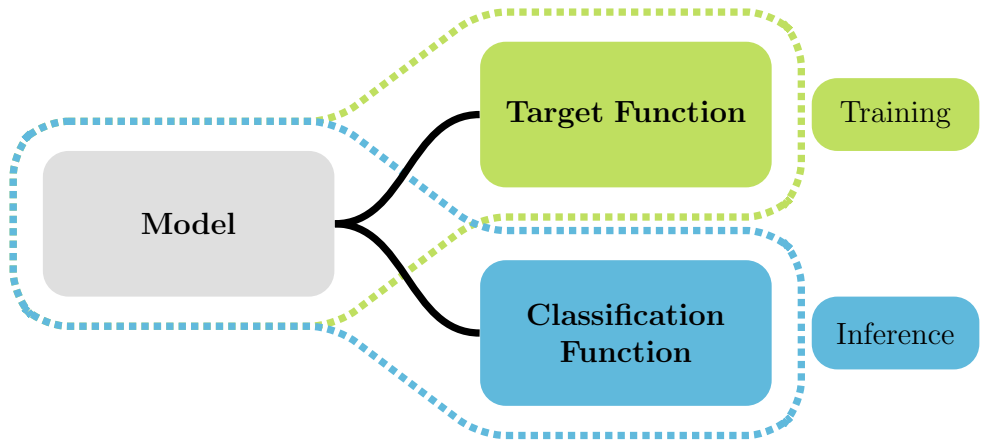

In this work, we combine different regression approaches into one framework.
To achieve this, we break the regression process into three exchangeable components, as shown in Figure \ref{fig:RegFramework}.

First, we need to define a model. 
The task of the model is to learn how to process the data. 
For our task of image processing, computer vision models are particularly useful. 
The important condition is that the output of the model can be any vector. 
Deep learning models are particularly suitable for this since they can be trained to map arbitrary inputs to arbitrary outputs.

Next, we define a target function. 
The purpose of the target function is to define the output on which the model will be trained.
This function is only needed for training and can be discarded after the model is fully trained.

The third step is to define a classification function.
The task of the classification function is to assign the output of the model to one of the possible classes, given a model that has already been trained on a certain target function.
Since the output vector of the network can take arbitrary values, the classification function must also be able to map vectors with arbitrary values to the corresponding classes. 
Since the interpretation of the output vector depends strongly on the target function on which the model was trained, not every classification function is a reasonable choice for every target function.
This relationship between the components can also be described as mappings.
The model can be thought of as a function
\begin{equation}
    f: \mathcal{I} \rightarrow \mathbb{R}^d .
\end{equation}
It maps each input in $\mathcal{I}$ to a vector in $\mathbb{R}^d$. 
The length $d$ of this vector is defined by the number of entries in the traget vector.
The target function $t$ can be defined as 
\begin{equation}
    t: \mathcal{K} \rightarrow \mathbb{R}^d .
\end{equation}
It maps each class in $\mathcal{K}$ to the target vector in $\mathbb{R}^d$.
The classification function
\begin{equation}
    c: \mathbb{R}^d \rightarrow \mathcal{K}
\end{equation}
maps an output of the model in $\mathbb{R}^d$ to one of the classes in $\mathcal{K}$.

\subsection{The Model}
We employ a ResNet50 \citep{He2016ResNet} and a Vision Transformer ViT-B-16 \citep{Dosovitskiy2021ViT}.
The ResNet is one of the most used CNN architectures.
In recent years, a new paradigm called Vision Transformers has emerged in the field of computer vision. 
Unlike architectures that rely solely on convolutions, Vision Transformers use the self-attention mechanism \citep{Vaswani2017} to process images.
Both models receive images as input and can output arbitrary vectors.
We chose these two models because they each represent a class of widely used computer vision architectures.
For the ResNet50, we replaced the last linear layer with an adapted one with the correct output size depending on the encoding used.
The last linear layer of the ViT-B-16, which calculates the output of the class token, was replaced in the same way.

\subsection{The Target Function}
In the following we describe the different types of target functions.
The encodings we use are not comprehensive, but they are the most commonly used.
A vector $\hat{y}_k$ refers to a target vector for label $k$, while a value $\hat{y}_{i,k}$ refers to the $i$-th entry in the target vector for label $k$.

\subsubsection{One-Hot Encoding}
The one-hot encoding maps each of the classes to a vector that has the value 1 in one entry and 0 in all others.
The index of the entry with value 1 corresponds to the ordinal class.
Here, for example, the class \textit{(+)} would be represented by:
\begin{equation}
    \hat{y}_2 = \left( \begin{array}{c} 0 \\ 1 \\ 0 \\ 0 \\ 0 \end{array} \right).
\end{equation}

\subsubsection{Gaussian Encoding}
A softer form of the one-hot encoding is the Gaussian encoding, which is also used by \citet{Tan2016}.
For an ordinal class $k$, the entries of the target vector $\hat{y}$ are defined by a Gaussian function as follows:
\begin{equation}
    \hat{y}_{i,k} = \frac{1}{\sqrt{2\pi\sigma^{2}}}\exp \left(-\frac{(i-k)^2}{2\sigma^2}\right).
\end{equation}
We set $\sigma^{2} = 1$ as suggested by \citet{Tan2016}.
Like the one-hot encoding, the Gaussian function has its maximum at the position whose index corresponds to the corresponding class.
However, the non-maximum values in the target vector are not 0, but decrease with increasing distance from the index position of the maximum value.
For example, for the class \textit{(+)}, the target vector would be given by:
\begin{equation}
    \hat{y}_2 = \left( \begin{array}{c} 0.242 \\ 0.399 \\ 0.242 \\ 0.054 \\ 0.004 \end{array} \right).
\end{equation}

\subsubsection{Continuous Encoding}
Another encoding is the equidistant mapping of ordinal labels to a continuous interval.
Here, the interval is chosen from 0 to 1, with the smallest class mapped to the value 0 and the largest class mapped to the value 1.
All other values in between are evenly distributed over the interval.
The target vector of this function contains only a single value, so it is equivalent to a scalar.
This value can be determined by:
\begin{equation}
    \hat{y}_k = \frac{k-1}{K-1}.
\end{equation}

\subsubsection{Progress-Bar Encoding}
Another approach is to encode the classes in a progress-bar, as described by \citet{Cheng2008Ord}.
We do not use a vector of length $K$, but as suggested by \citet{Niu2016}, a vector of length $K-1$.
The reason for this is that otherwise the first entry would be trained to always take the constant value 1 without exception, which would not add any information.
Mathematically, this can be defined as follows:
\begin{equation}
    \hat{y}_{i,k} = \left\{
    \begin{array}{ll}
        1, & \, \textrm{if}(i < k), \\
        0, & \, \textrm{otherwise.} \\
    \end{array}
    \right. 
\end{equation}

\subsubsection{Soft-Progress-Bar Encoding}
In addition, we propose an encoding derived from the progress-bar.
The soft-progress-bar works in a similar way to the progress-bar.
But between the transitions from 1 to 0, another value is inserted, which has the value 0.5.
This can be described by:
\begin{equation}
        \hat{y}_{i,k} = \left\{
        \begin{array}{ll}
            1, & \, \textrm{if}(i < k), \\
            0.5, & \, \textrm{if}(i = k), \\
            0, & \, \textrm{otherwise.} \\
        \end{array}
        \right. 
\end{equation}
The length of the vector is equal to the number of ordinal classes $K$.
The target vector for the class \textit{+} has the form:
\begin{equation}
    \hat{y}_3 = \left( \begin{array}{c} 1 \\ 1 \\ 0.5 \\ 0 \\ 0 \end{array} \right).
\end{equation}
The idea is that if the deviation is exactly one class, the difference in two entries will be 0.5.
However, if the classes differ by more than this, the differences will be larger.

\subsubsection{Binary Number Encoding}
A completely different encoding is the encoding of the class into binary numbers.
Here, the class $k$ is translated into a binary number, which is then encoded into a vector.
Thus, the vector for the class \textit{+}, would be translated into the binary representation 011 and thus into the vector
\begin{equation}
    \hat{y} = \left( \begin{array}{c} 0 \\ 1 \\ 1 \end{array} \right).
\end{equation}
In our case, the binary representation for the largest class $k=5$ has 3 entries, so the vector length here is 3.
The interesting thing about this encoding is that a single entry by itself tells us very little about the class.
For example, the last binary entry only tells us whether the class is odd or even in the ordinal ranking, but nothing more.
The situation is different with the progress-bar encoding. 
Based on the value of an entry, we can infer whether a certain threshold has been crossed or not.
This encoding was chosen not because we expect it to work well, but for completeness and to get a better idea of the impact of encoding on performance.

\subsection{The Classification Function}
Similar to the target function, many different classification functions can be defined.
Changing the classification function changes how the output of the trained model is interpreted.

A simple classification function is the argmax function.
It maps a vector to the index of the largest entry.
In our case, it is a reasonable choice only for the one-hot encoding and the Gaussian encoding.
For these target functions, the index of the largest entry is expected to be equal to the index of the class.

Since both the output of the model and the possible target vectors are vectors of the same length, it is possible to determine the distances between them and then select the class whose target vector has the smallest distance to the output vector.
For measuring the distances between vectors, $p$-norms are a possible choice.
Here we focus on the 1-norm (L1-distance), since it compares the differences element by element.
For an output vector, the L1 distance to a target vector of class $k$ with $n$ elements can be determined as follows:
\begin{equation}
    \norm{y-\hat{y}_k}_1 = \sum_{i=1}^{n}|y_i-\hat{y}_{i,k}|
\end{equation}

We compute the L1 distance for each of the possible target vectors to the output vector.
The selected class is the one for which the target vector is closest to the output vector.
This classification function can be used for all target functions described above.

Another metric for determining the distances between two vectors is the normalized dot product.
As with the L1 distance, the distance of the output vector $y$ to each of the target vectors $\hat{y}_k$ of class $k$ is then calculated:
\begin{equation}
    \frac{y \cdot \hat{y}_k}{\norm{y}_2\cdot \norm{\hat{y}_k}_2}.
\end{equation}
Here, the class of the target vector is taken which has the smallest distance to the output vector.
Since a scalar from the continuous encoding cannot be normalized, nor can the all-zero vector from the progress bar encoding, this classification function cannot be used with these target functions.

\subsection{Metrics}
To facilitate a meaningful comparison among various approaches, different forms of Cohen's kappa are employed in this context. 
Cohen's kappa serves as a metric to quantify the agreement between model predictions and ground truth.

The unweighted Cohen's kappa, was originally proposed by \citet{cohen1960coefficient}. 
It evaluates the exact match between the predicted class and the ground truth.
The proximity or distance between the classes is disregarded in this context. 
Expanding upon this foundation, the weighted Cohen's kappa \citep{Fleiss1969WeightedKappa}, also commonly known as Fleiss' kappa, was developed.
This variant considers deviations based on the degree of mismatch between the ground truth and the predictions.
For example, the linearly weighted kappa assigns linearly greater weight to deviations of greater magnitude.
A similar principle applies to the quadratic Cohen's kappa. 
Here, deviations are penalized with a quadratic weight factor.

Following the mathematical description \citep{Fleiss1969WeightedKappa} we define the proportion of occurences that $i$ is the true class and $j$ the predicted class as $p_{i,j}$. 
We define a weighting factor for each prediction-label pair $i,j$ as $w_{i,j}$. 
The weighting factors are set depending on whether the unweighted, linear, or quadratic Cohen's kappa is used.
From this we now calculate 
\begin{equation}
    p_o = \sum_{i=1}^{k}\sum_{j=1}^{k}w_{i,j}p_{i,j}
\end{equation}
and
\begin{equation}
    p_c = \sum_{i=1}^{k}\sum_{j=1}^{k}w_{i,j}p_{i,.}p_{.,j} .
\end{equation}
In the last step we then calculate the weighted kappa 
\begin{equation}
    k_w=\frac{p_o-p_c}{1-p_c}
\end{equation}

\section{Experimental Setup}
To evaluate and compare the approaches to ordinal regression, we use two well-established vision models. 
The choice encompasses the ResNet50, a widely recognized convolution-based architecture, and the ViT-B-16, a Transformer-based architecture. 
We choose these two architectures to explore the extent to which outcomes are influenced by architectural differences.

The implementation relies on PyTorch \citep{pytorch}. 
We train both a ResNet50 and a ViT-B-16.
For both architectures we use pre-trained versions provided by the PyTorch library.
We employ the mean squared error loss and the AdamW \citep{Loshchilov2019AdamW} optimizer.
Unlike cross-entropy loss, this works for all different encodings.
A consistent pipeline is used for both models. 
We train with a batch size of 32 for 30 epochs. 
The learning rate follows a cosine annealing schedule \citep{Loshchilov2017}.
The initial learning rate is set to $5\cdot 10^{-4}$ for the ResNet-50 and $5\cdot 10^{-5}$ for the ViT-B-16.
Different learning rates were tested for the different models and found to be good for the respective model.
Data augmentation is applied during training. 
We apply data augmentation similar to \citet{He2019}.
First, we randomly rotate the images within the range of $\pm 15^\circ$. 
Then we randomly stretch or compress them, varying the aspect ratio from $3/4$ to $4/3$. 
Next, we apply a random cropping operation to the images occupying an area from 8\% to 100\%, followed by a resizing operation to $224 \times 224$.
After that we apply color jittering by modifying the brightness with a factor of 0.4.
Our last step injects noise for each pixel and is drawn from a Gaussian distribution centered on the original value with a standard deviation of 0.1/255.
Here we use the same value as \citet{He2019} but only scaled since our image has values from 0 to one instead of 0 to 255.

Considering that x-ray images are gray scale only and lack the usual three color channels, we adopt the models to work with a single input channel.
To do this, we average the pre-trained weights across the three color channels, allowing them to be applied to a single-channel transformation.
The number of output values is equal to the length of an output or target vector multiplied by the number of different labels.
Here we have seven different labels.

The dataset is partitioned into a test dataset and a training dataset.
For this purpose, 20\% of the patients are used as a test set.
Subsequently, a five-fold cross-validation procedure is executed on the remaining training dataset.
For the evaluation, one-fifth of the training dataset is omitted in each iteration, while the remaining four-fifths are used for training. 
This process is repeated for the resulting five distinct training datasets. 
Care is taken to ensure that images of a given patient are exclusively present in a single portion of the dataset, i.e. we used patientwise stratification. 
The evaluation of the performance was always done on the same, from the training dataset distinct, test dataset.

The networks are trained separately for each target function and initialization is set to be identical across methods.
The evaluation after training has finished is performed on the independent test dataset that has not been seen during training. 
For the classification functions, the models are trained on the same data in combination with one of the classification functions.
In this context, the three different weighted Cohen's kappa coefficients are utilized.

\section{Results and Discussion}

\begin{table}[h!]
    \setlength{\tabcolsep}{2pt}
    \begin{center}
        \begin{tabular}{ l r | c c | c c | c c }
        Targ.-fn & Class.-fn & Unweight. $\kappa$ & \# & Lin. $\kappa$ & \# & Quad. $\kappa$ & \# \\
        \hline
        One-Hot & All & \textbf{0.420}\scriptsize{$\pm$1.09e-3} & 1 & 0.534\scriptsize{$\pm$1.09e-3} & 3 & 0.618\scriptsize{$\pm$1.15e-3} & 7 \\
        \hline
        Gauss & Argmax & 0.402\scriptsize{$\pm$1.67e-3} & 4 & 0.539\scriptsize{$\pm$1.19e-3} & 2 & 0.646\scriptsize{$\pm$9.58e-4} & 4 \\
        Gauss & L1 & 0.373\scriptsize{$\pm$1.12e-3} & 5 & 0.525\scriptsize{$\pm$8.60e-4} & 4 & 0.652\scriptsize{$\pm$6.75e-4} & 2 \\
        Gauss & DP & 0.365\scriptsize{$\pm$1.27e-3} & 7 & 0.520\scriptsize{$\pm$9.20e-4} & 6 & 0.651\scriptsize{$\pm$6.40e-4} & 3 \\
        \hline
        Prog-Bar & L1 & 0.410\scriptsize{$\pm$1.33e-3} & 2 & \textbf{0.541}\scriptsize{$\pm$1.24e-3} & 1 & 0.641\scriptsize{$\pm$1.20e-3} & 6 \\
        \hline
        Soft-Prog-Bar & L1 & 0.369\scriptsize{$\pm$9.25e-4} & 6 & 0.524\scriptsize{$\pm$7.61e-4} & 5 & \textbf{0.653}\scriptsize{$\pm$8.31e-4} & 1 \\
        Soft-Prog-Bar & DP & 0.331\scriptsize{$\pm$2.93e-3} & 10 & 0.479\scriptsize{$\pm$3.15e-3} & 9 & 0.607\scriptsize{$\pm$3.16e-3} & 8 \\
        \hline
        Continuous & L1 & 0.337\scriptsize{$\pm$9.28e-4} & 9 & 0.499\scriptsize{$\pm$1.10e-3} & 8 & 0.643\scriptsize{$\pm$1.41e-3} & 5 \\
        \hline
        Bin-Num & L1 & 0.404\scriptsize{$\pm$1.17e-3} & 3 & 0.505\scriptsize{$\pm$1.73e-3} & 7 & 0.579\scriptsize{$\pm$2.25e-3} & 9 \\
        Bin-Num & DP & 0.365\scriptsize{$\pm$2.40e-3} & 7 & 0.474\scriptsize{$\pm$3.06e-3} & 10 & 0.564\scriptsize{$\pm$3.34e-3} & 10 \\
        \end{tabular}
        \caption[TODO caption]{
                \label{tab:avgResNet50}
                The performance of the different methods using the ResNet50, measured with differently weighted Cohen's kappas.
                Here the average over all different diseases was used.
                The standard deviation (after ``$\pm$'') is calculated on the test set based on the five different trainings.
                The value \# represents the relative rank of the approach with respect to one weighting for the kappa.
        }
    \end{center}
    \vspace{1.0cm}
    \begin{center}
        \begin{tabular}{ l r | c c | c c | c c }
            Targ.-fn & Class.-fn & Unweight. $\kappa$ & \# & Lin. $\kappa$ & \# & Quad. $\kappa$ & \# \\
            \hline
            One-Hot & All & \textbf{0.386}\scriptsize{$\pm$1.65e-3} & 1 & 0.497\scriptsize{$\pm$1.58e-3} & 6 & 0.581\scriptsize{$\pm$1.52e-3} & 8 \\
            \hline
            Gauss & Argmax & 0.380\scriptsize{$\pm$1.21e-3} & 3 & \textbf{0.513}\scriptsize{$\pm$1.60e-3} & 1 & 0.617\scriptsize{$\pm$1.89e-3} & 4 \\
            Gauss & L1 & 0.355\scriptsize{$\pm$1.21e-3} & 5 & 0.502\scriptsize{$\pm$1.46e-3} & 3 & \textbf{0.624}\scriptsize{$\pm$1.93e-3} & 1 \\
            Gauss & DP & 0.349\scriptsize{$\pm$1.33e-3} & 7 & 0.498\scriptsize{$\pm$1.45e-3} & 4 & \textbf{0.624}\scriptsize{$\pm$1.90e-3} & 1 \\
            \hline
            Prog-Bar & L1 & 0.382\scriptsize{$\pm$1.16e-3} & 2 & 0.508\scriptsize{$\pm$8.74e-4} & 2 & 0.606\scriptsize{$\pm$8.82e-4} & 6 \\
            \hline
            Soft-Prog-Bar & L1 & 0.351\scriptsize{$\pm$1.16e-3} & 6 & 0.498\scriptsize{$\pm$1.73e-3} & 4 & 0.621\scriptsize{$\pm$2.19e-3} & 3 \\
            Soft-Prog-Bar & DP & 0.317\scriptsize{$\pm$8.72e-4} & 9 & 0.460\scriptsize{$\pm$8.54e-4} & 9 & 0.582\scriptsize{$\pm$9.12e-4} & 7 \\
            \hline
            Continuous & L1 & 0.313\scriptsize{$\pm$2.47e-3} & 10 & 0.471\scriptsize{$\pm$2.11e-3} & 7 & 0.611\scriptsize{$\pm$1.43e-3} & 5 \\
            \hline
            Bin-Num & L1 & 0.367\scriptsize{$\pm$1.07e-3} & 4 & 0.467\scriptsize{$\pm$1.24e-3} & 8 & 0.545\scriptsize{$\pm$1.25e-3} & 9 \\
            Bin-Num & DP & 0.344\scriptsize{$\pm$1.76e-3} & 8 & 0.448\scriptsize{$\pm$1.86e-3} & 10 & 0.535\scriptsize{$\pm$1.69e-3} & 10 \\
            \end{tabular}
        \caption[TODO caption]{
                \label{tab:avgViT}
                The performance of the different methods using the ViT-B-16, measured with differently weighted Cohen's kappas.
                Here the average over all different diseases was used.
                The standard deviation (after ``$\pm$'') is calculated on the test set based on the five different trainings.
                The value \# represents the relative rank of the approach with respect to one weighting for the kappa.
        }
    \end{center}
\end{table}

\begin{figure}[h!]
    \resizebox{\textwidth}{!}{
    \begin{tikzpicture}[
        one_hot_node/.style={rectangle, fill=cyan!70,minimum size=6mm},
        prog_bar_node/.style={rectangle, fill=orange!100,minimum size=6mm},
        bin_num_node/.style={rectangle, fill=violet!70,minimum size=6mm},
        gauss_node/.style={rectangle, fill=teal!70,minimum size=6mm},
        soft_prog_node/.style={rectangle, fill=olive!70,minimum size=6mm},
        continuous_node/.style={rectangle, fill=red!70,minimum size=6mm},
        one_hot_line/.style={draw=cyan!100,line width=1.0mm},
        prog_bar_line/.style={draw=orange!100,line width=1.0mm},
        bin_num_line/.style={draw=violet!100,line width=1.0mm},
        gauss_line/.style={draw=teal!100,line width=1.0mm},
        soft_prog_line/.style={draw=olive!100,line width=1.0mm},
        continuous_line/.style={draw=red!100,line width=1.0mm},
        ]
        
        \node[text width=3.6cm,align=center] () at (0,1.2) {\Large \textbf{Unweighted Kappa}};
        \node[one_hot_node,text width=3.6cm,align=right] (unw_One_Hot_All) at (0,0) {One-Hot-All};
        \node[prog_bar_node,text width=3.6cm,align=right] (unw_Prog_Bar_L1) at (0,-0.6) {Prog-Bar-L1};
        \node[bin_num_node,text width=3.6cm,align=right] (unw_Bin_Num_L1) at (0,-1.2) {Bin-Num-L1};
        \node[gauss_node,text width=3.6cm,align=right] (unw_Gauss_Argmax) at (0,-1.8) {Gauss-Argmax};
        \node[gauss_node,text width=3.6cm,align=right] (unw_Gauss_L1) at (0,-2.4) {Gauss-L1};
        \node[soft_prog_node,text width=3.6cm,align=right] (unw_Soft_Prog_Bar_L1) at (0,-3.0) {Soft-Prog-Bar-L1};
        \node[gauss_node,text width=3.6cm,align=right] (unw_Gauss_DP) at (0,-3.6) {Gauss-DP};
        \node[bin_num_node,text width=3.6cm,align=right] (unw_Bin_Num_DP) at (0,-4.2) {Bin-Num-DP};
        \node[continuous_node,text width=3.6cm,align=right] (unw_Continuous_L1) at (0,-4.8) {Continuous-L1};
        \node[soft_prog_node,text width=3.6cm,align=right] (unw_Soft_Prog_Bar_DP) at (0,-5.4) {Soft-Prog-Bar-DP};

        \node[text width=3.6cm,align=center] () at (7,1.2) {\Large \textbf{Linear Kappa}};
        \node[prog_bar_node,text width=3.6cm,align=center,] (lin_Prog_Bar_L1) at (7,0) {Prog-Bar-L1};
        \node[gauss_node,text width=3.6cm,align=center,] (lin_Gauss_Argmax) at (7,-0.6) {Gauss-Argmax};
        \node[one_hot_node,text width=3.6cm,align=center,] (lin_One_Hot_All) at (7,-1.2) {One-Hot-All};
        \node[gauss_node,text width=3.6cm,align=center,] (lin_Gauss_L1) at (7,-1.8) {Gauss-L1};
        \node[soft_prog_node,text width=3.6cm,align=center,] (lin_Soft_Prog_Bar_L1) at (7,-2.4) {Soft-Prog-Bar-L1};
        \node[gauss_node,text width=3.6cm,align=center,] (lin_Gauss_DP) at (7,-3.0) {Gauss-DP};
        \node[bin_num_node,text width=3.6cm,align=center,] (lin_Bin_Num_L1) at (7,-3.6) {Bin-Num-L1};
        \node[continuous_node,text width=3.6cm,align=center,] (lin_Continuous_L1) at (7,-4.2) {Continuous-L1};
        \node[soft_prog_node,text width=3.6cm,align=center,] (lin_Soft_Prog_Bar_DP) at (7,-4.8) {Soft-Prog-Bar-DP};
        \node[bin_num_node,text width=3.6cm,align=center,] (lin_Bin_Num_DP) at (7,-5.4) {Bin-Num-DP};

        \node[text width=3.6cm,align=center] () at (14,1.2) {\Large \textbf{Quadratic Kappa}};
        \node[soft_prog_node,text width=3.6cm,align=left] (quad_Soft_Prog_Bar_L1) at (14,0) {Soft-Prog-Bar-L1};
        \node[gauss_node,text width=3.6cm,align=left] (quad_Gauss_L1) at (14,-0.6) {Gauss-L1};
        \node[gauss_node,text width=3.6cm,align=left] (quad_Gauss_DP) at (14,-1.2) {Gauss-DP};
        \node[gauss_node,text width=3.6cm,align=left] (quad_Gauss_Argmax) at (14,-1.8) {Gauss-Argmax};
        \node[continuous_node,text width=3.6cm,align=left] (quad_Continuous_L1) at (14,-2.4) {Continuous-L1};
        \node[prog_bar_node,text width=3.6cm,align=left] (quad_Prog_Bar_L1) at (14,-3.0) {Prog-Bar-L1};
        \node[one_hot_node,text width=3.6cm,align=left] (quad_One_Hot_All) at (14,-3.6) {One-Hot-All};
        \node[soft_prog_node,text width=3.6cm,align=left] (quad_Soft_Prog_Bar_DP) at (14,-4.2) {Soft-Prog-Bar-DP};
        \node[bin_num_node,text width=3.6cm,align=left] (quad_Bin_Num_L1) at (14,-4.8) {Bin-Num-L1};
        \node[bin_num_node,text width=3.6cm,align=left] (quad_Bin_Num_DP) at (14,-5.4) {Bin-Num-DP};

        \draw[-,one_hot_line] (unw_One_Hot_All.east) -- (lin_One_Hot_All.west);
        \draw[-,prog_bar_line] (unw_Prog_Bar_L1.east) -- (lin_Prog_Bar_L1.west);
        \draw[-,bin_num_line] (unw_Bin_Num_L1.east) -- (lin_Bin_Num_L1.west);
        \draw[-,bin_num_line] (unw_Bin_Num_DP.east) -- (lin_Bin_Num_DP.west);
        \draw[-,gauss_line] (unw_Gauss_L1.east) -- (lin_Gauss_L1.west);
        \draw[-,gauss_line] (unw_Gauss_DP.east) -- (lin_Gauss_DP.west);
        \draw[-,gauss_line] (unw_Gauss_Argmax.east) -- (lin_Gauss_Argmax.west);
        \draw[-,soft_prog_line] (unw_Soft_Prog_Bar_L1.east) -- (lin_Soft_Prog_Bar_L1.west);
        \draw[-,soft_prog_line] (unw_Soft_Prog_Bar_DP.east) -- (lin_Soft_Prog_Bar_DP.west);
        \draw[-,continuous_line] (unw_Continuous_L1.east) -- (lin_Continuous_L1.west);

        \draw[-,one_hot_line] (lin_One_Hot_All.east) -- (quad_One_Hot_All.west);
        \draw[-,prog_bar_line] (lin_Prog_Bar_L1.east) -- (quad_Prog_Bar_L1.west);
        \draw[-,bin_num_line] (lin_Bin_Num_L1.east) -- (quad_Bin_Num_L1.west);
        \draw[-,bin_num_line] (lin_Bin_Num_DP.east) -- (quad_Bin_Num_DP.west);
        \draw[-,gauss_line] (lin_Gauss_L1.east) -- (quad_Gauss_L1.west);
        \draw[-,gauss_line] (lin_Gauss_DP.east) -- (quad_Gauss_DP.west);
        \draw[-,gauss_line] (lin_Gauss_Argmax.east) -- (quad_Gauss_Argmax.west);
        \draw[-,soft_prog_line] (lin_Soft_Prog_Bar_L1.east) -- (quad_Soft_Prog_Bar_L1.west);
        \draw[-,soft_prog_line] (lin_Soft_Prog_Bar_DP.east) -- (quad_Soft_Prog_Bar_DP.west);
        \draw[-,continuous_line] (lin_Continuous_L1.east) -- (quad_Continuous_L1.west);

    \end{tikzpicture}
}
    \caption{\label{fig:OrderResNet}
    The change of the ranking of the best methods of the ResNet50.
    It is possible to see which methods are better and which are worse, and which become worse when the weighting used for the kappa is changed.}
    \vspace{2cm}
    \resizebox{\textwidth}{!}{
    \begin{tikzpicture}[
        one_hot_node/.style={rectangle, fill=cyan!70,minimum size=6mm},
        prog_bar_node/.style={rectangle, fill=orange!100,minimum size=6mm},
        bin_num_node/.style={rectangle, fill=violet!70,minimum size=6mm},
        gauss_node/.style={rectangle, fill=teal!70,minimum size=6mm},
        soft_prog_node/.style={rectangle, fill=olive!70,minimum size=6mm},
        continuous_node/.style={rectangle, fill=red!70,minimum size=6mm},
        one_hot_line/.style={draw=cyan!100,line width=1.0mm},
        prog_bar_line/.style={draw=orange!100,line width=1.0mm},
        bin_num_line/.style={draw=violet!100,line width=1.0mm},
        gauss_line/.style={draw=teal!100,line width=1.0mm},
        soft_prog_line/.style={draw=olive!100,line width=1.0mm},
        continuous_line/.style={draw=red!100,line width=1.0mm},
        ]

        \node[text width=3.6cm,align=center] () at (0,1.2) {\Large \textbf{Unweighted Kappa}};
        \node[one_hot_node,text width=3.6cm,align=right] (unw_One_Hot_All) at (0,0) {One-Hot-All};
        \node[prog_bar_node,text width=3.6cm,align=right] (unw_Prog_Bar_L1) at (0,-0.6) {Prog-Bar-L1};
        \node[gauss_node,text width=3.6cm,align=right] (unw_Gauss_Argmax) at (0,-1.2) {Gauss-Argmax};
        \node[bin_num_node,text width=3.6cm,align=right] (unw_Bin_Num_L1) at (0,-1.8) {Bin-Num-L1};
        \node[gauss_node,text width=3.6cm,align=right] (unw_Gauss_L1) at (0,-2.4) {Gauss-L1};
        \node[soft_prog_node,text width=3.6cm,align=right] (unw_Soft_Prog_Bar_L1) at (0,-3.0) {Soft-Prog-Bar-L1};
        \node[gauss_node,text width=3.6cm,align=right] (unw_Gauss_DP) at (0,-3.6) {Gauss-DP};
        \node[bin_num_node,text width=3.6cm,align=right] (unw_Bin_Num_DP) at (0,-4.2) {Bin-Num-DP};
        \node[continuous_node,text width=3.6cm,align=right] (unw_Continuous_L1) at (0,-4.8) {Continuous-L1};
        \node[soft_prog_node,text width=3.6cm,align=right] (unw_Soft_Prog_Bar_DP) at (0,-5.4) {Soft-Prog-Bar-DP};

        \node[text width=3.6cm,align=center] () at (7,1.2) {\Large \textbf{Linear Kappa}};
        \node[gauss_node,text width=3.6cm,align=center,] (lin_Gauss_Argmax) at (7,0) {Gauss-Argmax};
        \node[prog_bar_node,text width=3.6cm,align=center,] (lin_Prog_Bar_L1) at (7,-0.6) {Prog-Bar-L1};
        \node[gauss_node,text width=3.6cm,align=center,] (lin_Gauss_L1) at (7,-1.2) {Gauss-L1};
        \node[gauss_node,text width=3.6cm,align=center,] (lin_Gauss_DP) at (7,-1.8) {Gauss-DP};
        \node[soft_prog_node,text width=3.6cm,align=center,] (lin_Soft_Prog_Bar_L1) at (7,-2.4) {Soft-Prog-Bar-L1};
        \node[one_hot_node,text width=3.6cm,align=center,] (lin_One_Hot_All) at (7,-3.0) {One-Hot-All};
        \node[continuous_node,text width=3.6cm,align=center,] (lin_Continuous_L1) at (7,-3.6) {Continuous-L1};
        \node[bin_num_node,text width=3.6cm,align=center,] (lin_Bin_Num_L1) at (7,-4.2) {Bin-Num-L1};
        \node[soft_prog_node,text width=3.6cm,align=center,] (lin_Soft_Prog_Bar_DP) at (7,-4.8) {Soft-Prog-Bar-DP};
        \node[bin_num_node,text width=3.6cm,align=center,] (lin_Bin_Num_DP) at (7,-5.4) {Bin-Num-DP};

        \node[text width=3.6cm,align=center] () at (14,1.2) {\Large \textbf{Quadratic Kappa}};
        \node[gauss_node,text width=3.6cm,align=left] (quad_Gauss_L1) at (14,0) {Gauss-L1};
        \node[gauss_node,text width=3.6cm,align=left] (quad_Gauss_DP) at (14,-0.6) {Gauss-DP};
        \node[soft_prog_node,text width=3.6cm,align=left] (quad_Soft_Prog_Bar_L1) at (14,-1.2) {Soft-Prog-Bar-L1};
        \node[gauss_node,text width=3.6cm,align=left] (quad_Gauss_Argmax) at (14,-1.8) {Gauss-Argmax};
        \node[continuous_node,text width=3.6cm,align=left] (quad_Continuous_L1) at (14,-2.4) {Continuous-L1};
        \node[prog_bar_node,text width=3.6cm,align=left] (quad_Prog_Bar_L1) at (14,-3.0) {Prog-Bar-L1};
        \node[one_hot_node,text width=3.6cm,align=left] (quad_One_Hot_All) at (14,-3.6) {One-Hot-All};
        \node[soft_prog_node,text width=3.6cm,align=left] (quad_Soft_Prog_Bar_DP) at (14,-4.2) {Soft-Prog-Bar-DP};
        \node[bin_num_node,text width=3.6cm,align=left] (quad_Bin_Num_L1) at (14,-4.8) {Bin-Num-L1};
        \node[bin_num_node,text width=3.6cm,align=left] (quad_Bin_Num_DP) at (14,-5.4) {Bin-Num-DP};

        \draw[-,one_hot_line] (unw_One_Hot_All.east) -- (lin_One_Hot_All.west);
        \draw[-,prog_bar_line] (unw_Prog_Bar_L1.east) -- (lin_Prog_Bar_L1.west);
        \draw[-,bin_num_line] (unw_Bin_Num_L1.east) -- (lin_Bin_Num_L1.west);
        \draw[-,bin_num_line] (unw_Bin_Num_DP.east) -- (lin_Bin_Num_DP.west);
        \draw[-,gauss_line] (unw_Gauss_L1.east) -- (lin_Gauss_L1.west);
        \draw[-,gauss_line] (unw_Gauss_DP.east) -- (lin_Gauss_DP.west);
        \draw[-,gauss_line] (unw_Gauss_Argmax.east) -- (lin_Gauss_Argmax.west);
        \draw[-,soft_prog_line] (unw_Soft_Prog_Bar_L1.east) -- (lin_Soft_Prog_Bar_L1.west);
        \draw[-,soft_prog_line] (unw_Soft_Prog_Bar_DP.east) -- (lin_Soft_Prog_Bar_DP.west);
        \draw[-,continuous_line] (unw_Continuous_L1.east) -- (lin_Continuous_L1.west);

        \draw[-,one_hot_line] (lin_One_Hot_All.east) -- (quad_One_Hot_All.west);
        \draw[-,prog_bar_line] (lin_Prog_Bar_L1.east) -- (quad_Prog_Bar_L1.west);
        \draw[-,bin_num_line] (lin_Bin_Num_L1.east) -- (quad_Bin_Num_L1.west);
        \draw[-,bin_num_line] (lin_Bin_Num_DP.east) -- (quad_Bin_Num_DP.west);
        \draw[-,gauss_line] (lin_Gauss_L1.east) -- (quad_Gauss_L1.west);
        \draw[-,gauss_line] (lin_Gauss_DP.east) -- (quad_Gauss_DP.west);
        \draw[-,gauss_line] (lin_Gauss_Argmax.east) -- (quad_Gauss_Argmax.west);
        \draw[-,soft_prog_line] (lin_Soft_Prog_Bar_L1.east) -- (quad_Soft_Prog_Bar_L1.west);
        \draw[-,soft_prog_line] (lin_Soft_Prog_Bar_DP.east) -- (quad_Soft_Prog_Bar_DP.west);
        \draw[-,continuous_line] (lin_Continuous_L1.east) -- (quad_Continuous_L1.west);

    \end{tikzpicture}
}
    \caption{\label{fig:OrderViT}The change of the ranking of the best methods of the ViT-B-16.
    It is clear that the order is different from the ResNet50 in Figure \ref{fig:OrderResNet}, but the changes are not too large. Good methods are still good, and bad methods are still bad.}
\end{figure}

Table \ref{tab:avgResNet50} and Table \ref{tab:avgViT} present the aggregated mean performance across all diseases as measured by the unweighted, the linearly weighted, and the quadratically weighted kappa coefficient. 
Since the one-hot encoding method produces the same results for all three classification functions, a unified representation of the three classification functions is used.
Thus, the label ``All'' refers to ``Argmax'', ``L1 Distance'', and ``Dot-Product'' in the context of one-hot encoding.
Figure \ref{fig:OrderResNet} and Figure \ref{fig:OrderViT} illustrate how the order of the different methods changes when measured with unweighted, linear, and quadratic Cohen's kappa. 
In the following, we investigate in more detail how the methods behave when using the different Cohen's kappas.

\subsection{Unweighted Cohen's Kappa}
When evaluating the unweighted Cohen's kappa, the one-hot encoding method outperforms all other methods for both the ResNet50 and ViT-B-16 architectures.
The focus of the one-hot encoding on the exact class match is consistent with the nature of the unweighted Cohen's kappa. 
In particular, the proximity or distance of classes is not considered, which is well reflected by this metric. 
Investigating the individual disease-specific scores (see appendix), the one-hot encoding appears at the top of the rankings in 13 out of 16 cases under the unweighted kappa.
However, the unweighted Cohen's kappa does not adequately reflect the importance of different misclassifications.
The exact labels are not highly relevant for our application, since the labels are based on the radiologist's judgment and may vary due to inter-reader variability.
Rather, it is more important that the deviation in predicting the correct ordinal class is not too large from the radiologist's rating.
This is not reflected in the unweighted kappa, but it is reflected in the weighted kappa.
This leads to the conclusion that although one-hot encoding performs best when using the unweighted Cohen's kappa, it is not the method of choice for our purpose because it does not perform best when using the weighted kappa.

\subsection{Linear Cohen's Kappa}
The linear Cohen's kappa better reflects the relevance of prediction errors to clinical practice. 
Because larger deviations are penalized more severely by the linear metric, this better reflects the impact of errors in the clinical setting than the unweighted Cohen's kappa, which penalizes all errors equally regardless of their size.
Interestingly, the order of the methods changes when the weighting of Cohen's kappa is changed, as can be seen in Figures \ref{fig:OrderResNet} and \ref{fig:OrderViT}.
This means that some methods, such as one-hot encoding, hit the correct class more often, but also miss by a larger margin when a deviation occurs than, for example, the progress-bar encoding.
Therefore, when evaluating with the linearly weighted Cohen's kappa, other methods prove to be particularly suitable.
In particular, the Gaussian encoding with argmax and the progress-bar interchangeably secure the first and second positions in the ranking. 
This pattern extends to the evaluation of individual diseases. 
Interestingly, the ResNet tends to perform better with the progress-bar encoding, while the ViT performs better with the Gaussian encoding in combination with argmax.
This observation holds not only for the average, but also for most of the diseases examined here.
This shows that different methods may be more appropriate for different models.
Comparing the ranking between the use of a ViT-B-16 and the use of a ResNet50 shows that the model used has little effect on the ranking of the encodings used.
A method that excels on one model remains effective on the other. 
These results highlight the lack of a universal regression method that is optimal for all problem settings.
However, depending on the application, it is also possible that a linear weighting of the errors still penalizes the deviations too little.
In such a case, the quadratically weighted kappa can be used.
This shifts the focus even more from hitting the exact class to avoiding very large outliers.

\subsection{Quadratic Cohen's Kappa}
Just as there are differences in the ranking of methods when using linearly weighted kappa instead of unweighted kappa, there are also differences between linearly and quadratically weighted kappa.
Once again, other methods have a better performance.
For ResNet50, the soft-progress-bar leads by a narrow margin, followed closely by Gaussian encoding with L1 distance and dot product.
In the context of ViT-B-16, the Gaussian encoding methods take the lead, with the soft-progress-bar and the L1 distance in second and third place, respectively.
Despite the variation in optimal methods, there is a remarkable consistency: a method that performs optimally for one model retains its effectiveness as one of the top choices for the other.

\subsection{Identifying Inadequate Methods}
Some of the encodings investigated show poor performance across all weightings of Cohen's Kappa.
In particular, the continuous encoding performs poorly, especially under unweighted and linearly weighted kappa. 
Even under quadratic weighted kappa, it ranks only fifth. 
Similarly, binary-number encoding performs poorly, with the exception of L1 distance under unweighted kappa.
But even here it only ranks third or fourth.
This may be due to the fact that binary encoding does not encode ordinality in a natural way.
The value of just one of the entries has very little meaning without knowledge of the other values.
For example, if the last entry takes the value one, this can represent any class with an odd ordinal rank.
For comparison, in the progress-bar encoding, each entry can be interpreted as a binary classification for a given threshold.
It is interesting to note that in most cases the L1 distance outperforms the normalized dot product. 
For this reason, we recommend using the L1 distance when running a regression on similar problems.
It is noteworthy that continuous regression, which is a very natural way to encode a regression problem, does not perform as well as the other methods.
In fact, it is one of the worst performing methods for unweighted Cohen's kappa.
Similarly, for the linear or quadratic kappa it does not perform as well as the other methods.
This suggests that the more complex structure of a coding such as the Gaussian coding is helpful for the task of ordinal regression.

\subsection{Magnitude of Performance Differences}
Comparing the best and worst performing methods reveals large discrepancies.
For instance, considering the ResNet50 results in Table \ref{tab:avgResNet50}, the performance gaps for various kappas between the first and last places are as follows:
unweighted kappa: 0.089; 
linear kappa: 0.067; 
quadratic kappa: 0.089. 
The discrepancies are similar for the ViT-B-16:
unweighted kappa: 0.073;
linear kappa: 0.065;
quadratic kappa: 0.089;
These variations underscore the significant impact of regression method selection on performance.
This shows that choosing the right ordinal regression method is important, and selecting the wrong one can result in a major performance loss.

\subsection{Practical Implications}
The results of this work have direct implications for the training of neural networks for such ordinal regression tasks.
Key practical tips based on these findings can be summarized as follows:
Choose a metric that best reflects the problem. 
The unweighted kappa is suitable for classification tasks; in such cases, use one-hot encoding.
For weighted deviations based on magnitude, other encodings are advisable. 
The framework presented here facilitates efficient implementation of various methods to determine the most suitable one for a specific application.
Whether the linear or the quadratic weighted kappa is better suited to evaluate the different possible estimation errors depends on the direct application.
Therefore, a thoughtful decision should be made about the most appropriate method based on the size and nature of the deviations before using ordinal regression.
It is also interesting to note that although ResNet50 and ViT-B-16 are fundamentally different architectures, the order of the methods is not substantially different.
However, in general, the ResNet50 tends to perform better on our dataset than the ViT-B-16.

\section{Conclusion}
In this study, we have combined various ordinal regression methodologies from the literature into a unified framework. 
The regression task was divided into three distinct parts:
The model trained on the task, the target function that outputs a vector for each class the model is trained on, and the classification function that maps the output of the already trained model to one of the possible classes.
These efforts have enabled a systematic comparison of the different methodologies to be made.
In order to emphasise different degrees of penalizing deviation, we used differently weighted Cohen's kappa coefficients for the measurement.
Our analysis showed: There is no single optimal method that is universally superior across all metrics used.
Subtle differences also emerged between the two different network architectures, showing consistency across different diseases.
In general, however, the one-hot encoding scheme proved to be the most effective for class-wise classification, as reflected by the unweighted Cohen's kappa. 
However, if the emphasis is on minimizing variation, Gaussian, progress-bar, or soft-progress-bar encoding should be preferred over the use of one-hot encoding.

\section*{Acknowledgements}
Computations were performed with computing resources granted by RWTH Aachen University under project rwth1444.

\pagebreak
\appendix
\section{All Results}
Results for the different diseases.
Comparison of these tables shows that performance is highly dependent on the disease.
Nevertheless, the order of the different combinations of target function and classification function remains roughly the same across the different diseases.

\subsection{Atelectasis Left}

\begin{table}[h]
    \setlength{\tabcolsep}{2pt}
    \begin{center}
        \begin{tabular}{ l r | c c | c c | c c }
        Targ.-fn & Class.-fn & Unweight. $\kappa$ & \# & Lin. $\kappa$ & \# & Quad. $\kappa$ & \# \\
        \hline
        One-Hot & All & \textbf{0.409}\scriptsize{$\pm$7.72e-4} & 1 & 0.492\scriptsize{$\pm$8.28e-4} & 2 & 0.554\scriptsize{$\pm$9.13e-4} & 7 \\
        \hline
        Gauss & Argmax & 0.379\scriptsize{$\pm$2.76e-3} & 4 & 0.492\scriptsize{$\pm$1.06e-3} & 2 & 0.579\scriptsize{$\pm$8.20e-4} & 4 \\
        Gauss & L1 & 0.339\scriptsize{$\pm$1.98e-3} & 6 & 0.473\scriptsize{$\pm$1.60e-3} & 4 & 0.585\scriptsize{$\pm$1.10e-3} & 2 \\
        Gauss & DP & 0.331\scriptsize{$\pm$1.77e-3} & 8 & 0.467\scriptsize{$\pm$1.52e-3} & 6 & 0.583\scriptsize{$\pm$1.09e-3} & 3 \\
        \hline
        Prog-Bar & L1 & 0.392\scriptsize{$\pm$9.81e-4} & 3 & \textbf{0.494}\scriptsize{$\pm$6.70e-4} & 1 & 0.571\scriptsize{$\pm$7.34e-4} & 6 \\
        \hline
        Soft-Prog-Bar & L1 & 0.333\scriptsize{$\pm$2.78e-3} & 7 & 0.471\scriptsize{$\pm$2.49e-3} & 5 & \textbf{0.588}\scriptsize{$\pm$2.11e-3} & 1 \\
        Soft-Prog-Bar & DP & 0.306\scriptsize{$\pm$3.90e-3} & 9 & 0.421\scriptsize{$\pm$4.59e-3} & 10 & 0.519\scriptsize{$\pm$5.06e-3} & 8 \\
        \hline
        Continuous & L1 & 0.289\scriptsize{$\pm$1.66e-3} & 10 & 0.437\scriptsize{$\pm$1.97e-3} & 8 & 0.574\scriptsize{$\pm$2.10e-3} & 5 \\
        \hline
        Bin-Num & L1 & 0.393\scriptsize{$\pm$3.44e-3} & 2 & 0.465\scriptsize{$\pm$4.18e-3} & 7 & 0.517\scriptsize{$\pm$5.16e-3} & 9 \\
        Bin-Num & DP & 0.343\scriptsize{$\pm$5.58e-3} & 5 & 0.427\scriptsize{$\pm$6.82e-3} & 9 & 0.496\scriptsize{$\pm$7.50e-3} & 10 \\
        \end{tabular}
        \caption[TODO caption]{
                \label{tab:atelectasisleftResNet50}atelectasis, left; ResNet50; \#: relative rank; value\scriptsize{$\pm$ SD}
        }
    \end{center}
\end{table}

\begin{table}[h]
    \setlength{\tabcolsep}{2pt}
    \begin{center}
        \begin{tabular}{ l r | c c | c c | c c }
        Targ.-fn & Class.-fn & Unweight. $\kappa$ & \# & Lin. $\kappa$ & \# & Quad. $\kappa$ & \# \\
        \hline
        One-Hot & All & \textbf{0.374}\scriptsize{$\pm$2.76e-3} & 1 & 0.454\scriptsize{$\pm$2.88e-3} & 3 & 0.515\scriptsize{$\pm$2.85e-3} & 7 \\
        \hline
        Gauss & Argmax & 0.356\scriptsize{$\pm$2.20e-3} & 4 & \textbf{0.463}\scriptsize{$\pm$1.83e-3} & 1 & 0.547\scriptsize{$\pm$2.71e-3} & 4 \\
        Gauss & L1 & 0.325\scriptsize{$\pm$1.87e-3} & 6 & 0.449\scriptsize{$\pm$1.72e-3} & 4 & \textbf{0.553}\scriptsize{$\pm$2.21e-3} & 1 \\
        Gauss & DP & 0.318\scriptsize{$\pm$2.22e-3} & 8 & 0.445\scriptsize{$\pm$1.92e-3} & 5 & 0.552\scriptsize{$\pm$2.18e-3} & 2 \\
        \hline
        Prog-Bar & L1 & 0.360\scriptsize{$\pm$4.28e-3} & 2 & 0.456\scriptsize{$\pm$4.00e-3} & 2 & 0.530\scriptsize{$\pm$3.76e-3} & 6 \\
        \hline
        Soft-Prog-Bar & L1 & 0.319\scriptsize{$\pm$1.74e-3} & 7 & 0.445\scriptsize{$\pm$2.45e-3} & 5 & 0.551\scriptsize{$\pm$3.52e-3} & 3 \\
        Soft-Prog-Bar & DP & 0.292\scriptsize{$\pm$2.20e-3} & 9 & 0.402\scriptsize{$\pm$2.08e-3} & 10 & 0.495\scriptsize{$\pm$2.32e-3} & 8 \\
        \hline
        Continuous & L1 & 0.274\scriptsize{$\pm$3.26e-3} & 10 & 0.413\scriptsize{$\pm$2.97e-3} & 8 & 0.541\scriptsize{$\pm$2.09e-3} & 5 \\
        \hline
        Bin-Num & L1 & 0.360\scriptsize{$\pm$2.04e-3} & 2 & 0.432\scriptsize{$\pm$1.75e-3} & 7 & 0.486\scriptsize{$\pm$2.10e-3} & 9 \\
        Bin-Num & DP & 0.328\scriptsize{$\pm$3.04e-3} & 5 & 0.407\scriptsize{$\pm$3.51e-3} & 9 & 0.472\scriptsize{$\pm$4.10e-3} & 10 \\
        \end{tabular}
        \caption[TODO caption]{
                \label{tab:atelectasisleftDeiT}atelectasis, left; ViT-B-16; \#: relative rank; value\scriptsize{$\pm$ SD}
        }
    \end{center}
\end{table}
\pagebreak

\subsection{Atelectasis Right}

\begin{table}[h]
    \setlength{\tabcolsep}{2pt}
    \begin{center}
        \begin{tabular}{ l r | c c | c c | c c }
        Targ.-fn & Class.-fn & Unweight. $\kappa$ & \# & Lin. $\kappa$ & \# & Quad. $\kappa$ & \# \\
        \hline
        One-Hot & All & \textbf{0.428}\scriptsize{$\pm$2.50e-3} & 1 & 0.521\scriptsize{$\pm$1.95e-3} & 3 & 0.591\scriptsize{$\pm$1.51e-3} & 7 \\
        \hline
        Gauss & Argmax & 0.400\scriptsize{$\pm$1.59e-3} & 4 & \textbf{0.523}\scriptsize{$\pm$9.01e-4} & 1 & 0.618\scriptsize{$\pm$2.02e-3} & 4 \\
        Gauss & L1 & 0.360\scriptsize{$\pm$3.28e-3} & 5 & 0.503\scriptsize{$\pm$2.13e-3} & 4 & 0.621\scriptsize{$\pm$9.53e-4} & 2 \\
        Gauss & DP & 0.350\scriptsize{$\pm$3.58e-3} & 8 & 0.497\scriptsize{$\pm$2.60e-3} & 6 & 0.620\scriptsize{$\pm$1.37e-3} & 3 \\
        \hline
        Prog-Bar & L1 & 0.411\scriptsize{$\pm$1.90e-3} & 3 & \textbf{0.523}\scriptsize{$\pm$1.31e-3} & 1 & 0.609\scriptsize{$\pm$1.72e-3} & 5 \\
        \hline
        Soft-Prog-Bar & L1 & 0.354\scriptsize{$\pm$2.22e-3} & 7 & 0.501\scriptsize{$\pm$1.79e-3} & 5 & \textbf{0.624}\scriptsize{$\pm$1.45e-3} & 1 \\
        Soft-Prog-Bar & DP & 0.310\scriptsize{$\pm$3.14e-3} & 9 & 0.442\scriptsize{$\pm$3.41e-3} & 10 & 0.558\scriptsize{$\pm$3.26e-3} & 8 \\
        \hline
        Continuous & L1 & 0.306\scriptsize{$\pm$1.82e-3} & 10 & 0.465\scriptsize{$\pm$1.80e-3} & 8 & 0.608\scriptsize{$\pm$2.02e-3} & 6 \\
        \hline
        Bin-Num & L1 & 0.414\scriptsize{$\pm$2.45e-3} & 2 & 0.494\scriptsize{$\pm$3.45e-3} & 7 & 0.552\scriptsize{$\pm$4.10e-3} & 9 \\
        Bin-Num & DP & 0.359\scriptsize{$\pm$6.83e-3} & 6 & 0.456\scriptsize{$\pm$8.03e-3} & 9 & 0.537\scriptsize{$\pm$8.15e-3} & 10 \\
        \end{tabular}
        \caption[TODO caption]{
                \label{tab:atelectasisrightResNet50}atelectasis, right; ResNet50; \#: relative rank; value\scriptsize{$\pm$ SD}
        }
    \end{center}
\end{table}

\begin{table}[h]
    \setlength{\tabcolsep}{2pt}
    \begin{center}
        \begin{tabular}{ l r | c c | c c | c c }
        Targ.-fn & Class.-fn & Unweight. $\kappa$ & \# & Lin. $\kappa$ & \# & Quad. $\kappa$ & \# \\
        \hline
        One-Hot & All & \textbf{0.392}\scriptsize{$\pm$1.52e-3} & 1 & 0.485\scriptsize{$\pm$1.73e-3} & 3 & 0.556\scriptsize{$\pm$2.20e-3} & 7 \\
        \hline
        Gauss & Argmax & 0.376\scriptsize{$\pm$3.91e-3} & 4 & \textbf{0.496}\scriptsize{$\pm$3.41e-3} & 1 & 0.589\scriptsize{$\pm$3.02e-3} & 4 \\
        Gauss & L1 & 0.345\scriptsize{$\pm$2.68e-3} & 6 & 0.482\scriptsize{$\pm$2.66e-3} & 4 & \textbf{0.595}\scriptsize{$\pm$2.92e-3} & 1 \\
        Gauss & DP & 0.337\scriptsize{$\pm$2.53e-3} & 8 & 0.477\scriptsize{$\pm$2.48e-3} & 6 & 0.594\scriptsize{$\pm$2.79e-3} & 2 \\
        \hline
        Prog-Bar & L1 & 0.382\scriptsize{$\pm$2.40e-3} & 2 & 0.490\scriptsize{$\pm$2.72e-3} & 2 & 0.574\scriptsize{$\pm$3.34e-3} & 6 \\
        \hline
        Soft-Prog-Bar & L1 & 0.340\scriptsize{$\pm$3.21e-3} & 7 & 0.478\scriptsize{$\pm$3.88e-3} & 5 & 0.592\scriptsize{$\pm$4.23e-3} & 3 \\
        Soft-Prog-Bar & DP & 0.300\scriptsize{$\pm$3.19e-3} & 9 & 0.425\scriptsize{$\pm$3.41e-3} & 10 & 0.534\scriptsize{$\pm$3.53e-3} & 8 \\
        \hline
        Continuous & L1 & 0.290\scriptsize{$\pm$2.99e-3} & 10 & 0.444\scriptsize{$\pm$2.07e-3} & 8 & 0.582\scriptsize{$\pm$1.56e-3} & 5 \\
        \hline
        Bin-Num & L1 & 0.381\scriptsize{$\pm$3.00e-3} & 3 & 0.462\scriptsize{$\pm$2.57e-3} & 7 & 0.524\scriptsize{$\pm$2.78e-3} & 9 \\
        Bin-Num & DP & 0.346\scriptsize{$\pm$2.73e-3} & 5 & 0.438\scriptsize{$\pm$2.49e-3} & 9 & 0.515\scriptsize{$\pm$2.41e-3} & 10 \\
        \end{tabular}
        \caption[TODO caption]{
                \label{tab:atelectasisrightDeiT}atelectasis, right; ViT-B-16; \#: relative rank; value\scriptsize{$\pm$ SD}
        }
    \end{center}
\end{table}
\pagebreak

\subsection{Congestion}

\begin{table}[h]
    \setlength{\tabcolsep}{2pt}
    \begin{center}
        \begin{tabular}{ l r | c c | c c | c c }
        Targ.-fn & Class.-fn & Unweight. $\kappa$ & \# & Lin. $\kappa$ & \# & Quad. $\kappa$ & \# \\
        \hline
        One-Hot & All & \textbf{0.324}\scriptsize{$\pm$1.29e-3} & 1 & 0.417\scriptsize{$\pm$7.86e-4} & 2 & 0.489\scriptsize{$\pm$5.53e-4} & 7 \\
        \hline
        Gauss & Argmax & 0.293\scriptsize{$\pm$3.10e-3} & 4 & 0.416\scriptsize{$\pm$1.46e-3} & 3 & \textbf{0.519}\scriptsize{$\pm$1.93e-3} & 1 \\
        Gauss & L1 & 0.252\scriptsize{$\pm$2.53e-3} & 5 & 0.389\scriptsize{$\pm$2.00e-3} & 4 & 0.515\scriptsize{$\pm$2.49e-3} & 3 \\
        Gauss & DP & 0.242\scriptsize{$\pm$2.36e-3} & 7 & 0.381\scriptsize{$\pm$2.08e-3} & 6 & 0.512\scriptsize{$\pm$2.96e-3} & 4 \\
        \hline
        Prog-Bar & L1 & 0.306\scriptsize{$\pm$2.18e-3} & 2 & \textbf{0.420}\scriptsize{$\pm$1.12e-3} & 1 & 0.512\scriptsize{$\pm$6.69e-4} & 4 \\
        \hline
        Soft-Prog-Bar & L1 & 0.250\scriptsize{$\pm$1.50e-3} & 6 & 0.389\scriptsize{$\pm$1.42e-3} & 4 & \textbf{0.519}\scriptsize{$\pm$1.36e-3} & 1 \\
        Soft-Prog-Bar & DP & 0.207\scriptsize{$\pm$3.88e-3} & 10 & 0.327\scriptsize{$\pm$4.52e-3} & 9 & 0.438\scriptsize{$\pm$5.18e-3} & 8 \\
        \hline
        Continuous & L1 & 0.212\scriptsize{$\pm$3.20e-3} & 9 & 0.354\scriptsize{$\pm$3.93e-3} & 8 & 0.498\scriptsize{$\pm$4.07e-3} & 6 \\
        \hline
        Bin-Num & L1 & 0.297\scriptsize{$\pm$3.10e-3} & 3 & 0.371\scriptsize{$\pm$5.00e-3} & 7 & 0.426\scriptsize{$\pm$6.95e-3} & 9 \\
        Bin-Num & DP & 0.240\scriptsize{$\pm$4.49e-3} & 8 & 0.322\scriptsize{$\pm$4.74e-3} & 10 & 0.399\scriptsize{$\pm$4.80e-3} & 10 \\
        \end{tabular}
        \caption[TODO caption]{
                \label{tab:congestionResNet50}congestion; ResNet50; \#: relative rank; value\scriptsize{$\pm$ SD}
        }
    \end{center}
\end{table}

\begin{table}[h]
    \setlength{\tabcolsep}{2pt}
    \begin{center}
        \begin{tabular}{ l r | c c | c c | c c }
        Targ.-fn & Class.-fn & Unweight. $\kappa$ & \# & Lin. $\kappa$ & \# & Quad. $\kappa$ & \# \\
        \hline
        One-Hot & All & \textbf{0.283}\scriptsize{$\pm$2.93e-3} & 1 & 0.369\scriptsize{$\pm$3.39e-3} & 3 & 0.436\scriptsize{$\pm$3.66e-3} & 7 \\
        \hline
        Gauss & Argmax & 0.276\scriptsize{$\pm$4.84e-3} & 3 & \textbf{0.388}\scriptsize{$\pm$5.46e-3} & 1 & 0.482\scriptsize{$\pm$6.06e-3} & 2 \\
        Gauss & L1 & 0.241\scriptsize{$\pm$3.67e-3} & 5 & 0.368\scriptsize{$\pm$4.89e-3} & 4 & \textbf{0.484}\scriptsize{$\pm$6.64e-3} & 1 \\
        Gauss & DP & 0.233\scriptsize{$\pm$4.14e-3} & 7 & 0.361\scriptsize{$\pm$5.08e-3} & 6 & 0.481\scriptsize{$\pm$6.75e-3} & 3 \\
        \hline
        Prog-Bar & L1 & 0.279\scriptsize{$\pm$2.38e-3} & 2 & 0.381\scriptsize{$\pm$2.04e-3} & 2 & 0.466\scriptsize{$\pm$1.72e-3} & 5 \\
        \hline
        Soft-Prog-Bar & L1 & 0.237\scriptsize{$\pm$1.04e-3} & 6 & 0.363\scriptsize{$\pm$1.68e-3} & 5 & 0.480\scriptsize{$\pm$2.50e-3} & 4 \\
        Soft-Prog-Bar & DP & 0.198\scriptsize{$\pm$1.98e-3} & 10 & 0.312\scriptsize{$\pm$9.40e-4} & 9 & 0.417\scriptsize{$\pm$6.72e-4} & 8 \\
        \hline
        Continuous & L1 & 0.200\scriptsize{$\pm$2.69e-3} & 9 & 0.330\scriptsize{$\pm$2.40e-3} & 7 & 0.461\scriptsize{$\pm$2.69e-3} & 6 \\
        \hline
        Bin-Num & L1 & 0.255\scriptsize{$\pm$3.64e-3} & 4 & 0.328\scriptsize{$\pm$4.42e-3} & 8 & 0.386\scriptsize{$\pm$4.59e-3} & 9 \\
        Bin-Num & DP & 0.226\scriptsize{$\pm$2.17e-3} & 8 & 0.304\scriptsize{$\pm$1.59e-3} & 10 & 0.376\scriptsize{$\pm$1.16e-3} & 10 \\
        \end{tabular}
        \caption[TODO caption]{
                \label{tab:congestionDeiT}congestion; ViT-B-16; \#: relative rank; value\scriptsize{$\pm$ SD}
        }
    \end{center}
\end{table}

\pagebreak
\subsection{Pleural Effusion Left}

\begin{table}[h]
    \setlength{\tabcolsep}{2pt}
    \begin{center}
        \begin{tabular}{ l r | c c | c c | c c }
        Targ.-fn & Class.-fn & Unweight. $\kappa$ & \# & Lin. $\kappa$ & \# & Quad. $\kappa$ & \# \\
        \hline
        One-Hot & All & \textbf{0.477}\scriptsize{$\pm$9.98e-4} & 1 & 0.587\scriptsize{$\pm$1.10e-3} & 3 & 0.666\scriptsize{$\pm$1.24e-3} & 7 \\
        \hline
        Gauss & Argmax & 0.461\scriptsize{$\pm$1.57e-3} & 3 & 0.595\scriptsize{$\pm$1.18e-3} & 2 & 0.701\scriptsize{$\pm$8.94e-4} & 3 \\
        Gauss & L1 & 0.438\scriptsize{$\pm$1.13e-3} & 5 & 0.582\scriptsize{$\pm$1.17e-3} & 4 & 0.702\scriptsize{$\pm$1.07e-3} & 2 \\
        Gauss & DP & 0.434\scriptsize{$\pm$1.04e-3} & 7 & 0.579\scriptsize{$\pm$9.95e-4} & 6 & 0.701\scriptsize{$\pm$7.70e-4} & 3 \\
        \hline
        Prog-Bar & L1 & 0.467\scriptsize{$\pm$1.69e-3} & 2 & \textbf{0.597}\scriptsize{$\pm$1.12e-3} & 1 & 0.697\scriptsize{$\pm$3.38e-3} & 5 \\
        \hline
        Soft-Prog-Bar & L1 & 0.437\scriptsize{$\pm$1.54e-3} & 6 & 0.582\scriptsize{$\pm$1.33e-3} & 4 & \textbf{0.704}\scriptsize{$\pm$1.36e-3} & 1 \\
        Soft-Prog-Bar & DP & 0.400\scriptsize{$\pm$2.20e-3} & 9 & 0.540\scriptsize{$\pm$2.69e-3} & 9 & 0.660\scriptsize{$\pm$3.52e-3} & 8 \\
        \hline
        Continuous & L1 & 0.411\scriptsize{$\pm$1.43e-3} & 8 & 0.562\scriptsize{$\pm$1.25e-3} & 7 & 0.696\scriptsize{$\pm$8.65e-4} & 6 \\
        \hline
        Bin-Num & L1 & 0.454\scriptsize{$\pm$2.06e-3} & 4 & 0.556\scriptsize{$\pm$1.83e-3} & 8 & 0.630\scriptsize{$\pm$1.52e-3} & 9 \\
        Bin-Num & DP & 0.395\scriptsize{$\pm$1.92e-3} & 10 & 0.496\scriptsize{$\pm$2.29e-3} & 10 & 0.578\scriptsize{$\pm$2.43e-3} & 10 \\
        \end{tabular}
        \caption[TODO caption]{
                \label{tab:pleuraleffusionleftResNet50}pleural effusion, left; ResNet50; \#: relative rank; value\scriptsize{$\pm$ SD}
        }
    \end{center}
\end{table}

\begin{table}[h]
    \setlength{\tabcolsep}{2pt}
    \begin{center}
        \begin{tabular}{ l r | c c | c c | c c }
        Targ.-fn & Class.-fn & Unweight. $\kappa$ & \# & Lin. $\kappa$ & \# & Quad. $\kappa$ & \# \\
        \hline
        One-Hot & All & \textbf{0.450}\scriptsize{$\pm$2.94e-3} & 1 & 0.561\scriptsize{$\pm$2.63e-3} & 4 & 0.644\scriptsize{$\pm$2.25e-3} & 8 \\
        \hline
        Gauss & Argmax & 0.439\scriptsize{$\pm$1.59e-3} & 2 & \textbf{0.572}\scriptsize{$\pm$1.34e-3} & 1 & 0.677\scriptsize{$\pm$1.82e-3} & 4 \\
        Gauss & L1 & 0.421\scriptsize{$\pm$2.01e-3} & 5 & 0.562\scriptsize{$\pm$1.05e-3} & 3 & \textbf{0.680}\scriptsize{$\pm$1.79e-3} & 1 \\
        Gauss & DP & 0.417\scriptsize{$\pm$2.21e-3} & 7 & 0.560\scriptsize{$\pm$1.49e-3} & 5 & \textbf{0.680}\scriptsize{$\pm$2.20e-3} & 1 \\
        \hline
        Prog-Bar & L1 & 0.439\scriptsize{$\pm$2.74e-3} & 2 & 0.568\scriptsize{$\pm$2.80e-3} & 2 & 0.670\scriptsize{$\pm$2.90e-3} & 5 \\
        \hline
        Soft-Prog-Bar & L1 & 0.419\scriptsize{$\pm$1.94e-3} & 6 & 0.560\scriptsize{$\pm$1.75e-3} & 5 & 0.678\scriptsize{$\pm$1.69e-3} & 3 \\
        Soft-Prog-Bar & DP & 0.391\scriptsize{$\pm$3.76e-3} & 8 & 0.528\scriptsize{$\pm$3.09e-3} & 8 & 0.645\scriptsize{$\pm$3.04e-3} & 7 \\
        \hline
        Continuous & L1 & 0.386\scriptsize{$\pm$3.82e-3} & 9 & 0.537\scriptsize{$\pm$3.48e-3} & 7 & 0.670\scriptsize{$\pm$3.51e-3} & 5 \\
        \hline
        Bin-Num & L1 & 0.426\scriptsize{$\pm$3.93e-3} & 4 & 0.528\scriptsize{$\pm$3.66e-3} & 8 & 0.604\scriptsize{$\pm$3.50e-3} & 9 \\
        Bin-Num & DP & 0.386\scriptsize{$\pm$6.51e-4} & 9 & 0.486\scriptsize{$\pm$1.12e-3} & 10 & 0.568\scriptsize{$\pm$1.58e-3} & 10 \\
        \end{tabular}
        \caption[TODO caption]{
                \label{tab:pleuraleffusionleftDeiT}pleural effusion, left; ViT-B-16; \#: relative rank; value\scriptsize{$\pm$ SD}
        }
    \end{center}
\end{table}

\pagebreak
\subsection{Pleural Effusion Right}

\begin{table}[h]
    \setlength{\tabcolsep}{2pt}
    \begin{center}
        \begin{tabular}{ l r | c c | c c | c c }
        Targ.-fn & Class.-fn & Unweight. $\kappa$ & \# & Lin. $\kappa$ & \# & Quad. $\kappa$ & \# \\
        \hline
        One-Hot & All & \textbf{0.532}\scriptsize{$\pm$1.60e-3} & 1 & 0.658\scriptsize{$\pm$1.43e-3} & 4 & 0.745\scriptsize{$\pm$1.45e-3} & 8 \\
        \hline
        Gauss & Argmax & 0.528\scriptsize{$\pm$1.41e-3} & 3 & \textbf{0.668}\scriptsize{$\pm$7.28e-4} & 1 & 0.768\scriptsize{$\pm$1.18e-3} & 5 \\
        Gauss & L1 & 0.507\scriptsize{$\pm$5.83e-4} & 5 & 0.660\scriptsize{$\pm$6.97e-4} & 3 & \textbf{0.775}\scriptsize{$\pm$1.01e-3} & 1 \\
        Gauss & DP & 0.503\scriptsize{$\pm$9.28e-4} & 7 & 0.657\scriptsize{$\pm$9.94e-4} & 6 & \textbf{0.775}\scriptsize{$\pm$1.14e-3} & 1 \\
        \hline
        Prog-Bar & L1 & 0.531\scriptsize{$\pm$1.79e-3} & 2 & \textbf{0.668}\scriptsize{$\pm$1.61e-3} & 1 & 0.766\scriptsize{$\pm$1.73e-3} & 6 \\
        \hline
        Soft-Prog-Bar & L1 & 0.505\scriptsize{$\pm$1.75e-3} & 6 & 0.658\scriptsize{$\pm$1.46e-3} & 4 & 0.774\scriptsize{$\pm$1.23e-3} & 3 \\
        Soft-Prog-Bar & DP & 0.465\scriptsize{$\pm$4.23e-3} & 10 & 0.625\scriptsize{$\pm$2.77e-3} & 9 & 0.752\scriptsize{$\pm$1.40e-3} & 7 \\
        \hline
        Continuous & L1 & 0.483\scriptsize{$\pm$1.33e-3} & 8 & 0.644\scriptsize{$\pm$1.15e-3} & 7 & 0.771\scriptsize{$\pm$1.03e-3} & 4 \\
        \hline
        Bin-Num & L1 & 0.518\scriptsize{$\pm$1.37e-3} & 4 & 0.636\scriptsize{$\pm$1.47e-3} & 8 & 0.718\scriptsize{$\pm$1.37e-3} & 9 \\
        Bin-Num & DP & 0.471\scriptsize{$\pm$1.90e-3} & 9 & 0.592\scriptsize{$\pm$2.17e-3} & 10 & 0.685\scriptsize{$\pm$2.15e-3} & 10 \\
        \end{tabular}
        \caption[TODO caption]{
                \label{tab:pleuraleffusionrightResNet50}pleural effusion, right; ResNet50; \#: relative rank; value\scriptsize{$\pm$ SD}
        }
    \end{center}
\end{table}

\begin{table}[h]
    \setlength{\tabcolsep}{2pt}
    \begin{center}
        \begin{tabular}{ l r | c c | c c | c c }
        Targ.-fn & Class.-fn & Unweight. $\kappa$ & \# & Lin. $\kappa$ & \# & Quad. $\kappa$ & \# \\
        \hline
        One-Hot & All & 0.503\scriptsize{$\pm$2.27e-3} & 3 & 0.633\scriptsize{$\pm$1.99e-3} & 6 & 0.725\scriptsize{$\pm$1.77e-3} & 8 \\
        \hline
        Gauss & Argmax & \textbf{0.506}\scriptsize{$\pm$1.92e-3} & 1 & \textbf{0.648}\scriptsize{$\pm$1.34e-3} & 1 & 0.751\scriptsize{$\pm$1.12e-3} & 4 \\
        Gauss & L1 & 0.485\scriptsize{$\pm$2.38e-3} & 4 & 0.639\scriptsize{$\pm$1.75e-3} & 3 & 0.756\scriptsize{$\pm$1.25e-3} & 2 \\
        Gauss & DP & 0.482\scriptsize{$\pm$2.99e-3} & 6 & 0.638\scriptsize{$\pm$2.06e-3} & 4 & \textbf{0.757}\scriptsize{$\pm$1.26e-3} & 1 \\
        \hline
        Prog-Bar & L1 & 0.504\scriptsize{$\pm$2.95e-3} & 2 & 0.644\scriptsize{$\pm$2.61e-3} & 2 & 0.745\scriptsize{$\pm$2.40e-3} & 5 \\
        \hline
        Soft-Prog-Bar & L1 & 0.482\scriptsize{$\pm$8.42e-4} & 6 & 0.637\scriptsize{$\pm$9.35e-4} & 5 & 0.755\scriptsize{$\pm$1.52e-3} & 3 \\
        Soft-Prog-Bar & DP & 0.447\scriptsize{$\pm$1.58e-3} & 9 & 0.607\scriptsize{$\pm$1.38e-3} & 8 & 0.734\scriptsize{$\pm$1.33e-3} & 7 \\
        \hline
        Continuous & L1 & 0.446\scriptsize{$\pm$2.97e-3} & 10 & 0.612\scriptsize{$\pm$2.62e-3} & 7 & 0.745\scriptsize{$\pm$1.96e-3} & 5 \\
        \hline
        Bin-Num & L1 & 0.483\scriptsize{$\pm$2.45e-3} & 5 & 0.601\scriptsize{$\pm$2.13e-3} & 9 & 0.686\scriptsize{$\pm$2.68e-3} & 9 \\
        Bin-Num & DP & 0.451\scriptsize{$\pm$2.49e-3} & 8 & 0.570\scriptsize{$\pm$2.91e-3} & 10 & 0.661\scriptsize{$\pm$3.14e-3} & 10 \\
        \end{tabular}
        \caption[TODO caption]{
                \label{tab:pleuraleffusionrightDeiT}pleural effusion, right; ViT-B-16; \#: relative rank; value\scriptsize{$\pm$ SD}
        }
    \end{center}
\end{table}

\pagebreak
\subsection{Pneumonic Infiltrates Left}

\begin{table}[h]
    \setlength{\tabcolsep}{2pt}
    \begin{center}
        \begin{tabular}{ l r | c c | c c | c c }
        Targ.-fn & Class.-fn & Unweight. $\kappa$ & \# & Lin. $\kappa$ & \# & Quad. $\kappa$ & \# \\
        \hline
        One-Hot & All & 0.370\scriptsize{$\pm$3.55e-3} & 2 & 0.514\scriptsize{$\pm$4.17e-3} & 6 & 0.622\scriptsize{$\pm$4.69e-3} & 8 \\
        \hline
        Gauss & Argmax & 0.367\scriptsize{$\pm$4.63e-3} & 3 & 0.526\scriptsize{$\pm$5.49e-3} & 2 & 0.652\scriptsize{$\pm$5.49e-3} & 5 \\
        Gauss & L1 & 0.351\scriptsize{$\pm$2.72e-3} & 6 & 0.523\scriptsize{$\pm$3.70e-3} & 3 & 0.665\scriptsize{$\pm$4.31e-3} & 2 \\
        Gauss & DP & 0.340\scriptsize{$\pm$2.44e-3} & 8 & 0.517\scriptsize{$\pm$3.05e-3} & 5 & \textbf{0.668}\scriptsize{$\pm$3.72e-3} & 1 \\
        \hline
        Prog-Bar & L1 & \textbf{0.371}\scriptsize{$\pm$3.12e-3} & 1 & \textbf{0.528}\scriptsize{$\pm$3.42e-3} & 1 & 0.650\scriptsize{$\pm$3.42e-3} & 7 \\
        \hline
        Soft-Prog-Bar & L1 & 0.348\scriptsize{$\pm$3.47e-3} & 7 & 0.520\scriptsize{$\pm$4.55e-3} & 4 & 0.664\scriptsize{$\pm$4.99e-3} & 3 \\
        Soft-Prog-Bar & DP & 0.305\scriptsize{$\pm$3.68e-3} & 10 & 0.488\scriptsize{$\pm$3.89e-3} & 9 & 0.652\scriptsize{$\pm$3.84e-3} & 5 \\
        \hline
        Continuous & L1 & 0.320\scriptsize{$\pm$2.53e-3} & 9 & 0.501\scriptsize{$\pm$3.56e-3} & 8 & 0.662\scriptsize{$\pm$4.59e-3} & 4 \\
        \hline
        Bin-Num & L1 & 0.359\scriptsize{$\pm$2.80e-3} & 5 & 0.488\scriptsize{$\pm$2.66e-3} & 9 & 0.586\scriptsize{$\pm$2.79e-3} & 10 \\
        Bin-Num & DP & 0.365\scriptsize{$\pm$2.94e-3} & 4 & 0.503\scriptsize{$\pm$2.29e-3} & 7 & 0.616\scriptsize{$\pm$2.07e-3} & 9 \\
        \end{tabular}
        \caption[TODO caption]{
                \label{tab:pneumonicinfiltratesleftResNet50}pneumonic infiltrates, left; ResNet50; \#: relative rank; value\scriptsize{$\pm$ SD}
        }
    \end{center}
\end{table}

\begin{table}[h]
    \setlength{\tabcolsep}{2pt}
    \begin{center}
        \begin{tabular}{ l r | c c | c c | c c }
        Targ.-fn & Class.-fn & Unweight. $\kappa$ & \# & Lin. $\kappa$ & \# & Quad. $\kappa$ & \# \\
        \hline
        One-Hot & All & 0.332\scriptsize{$\pm$3.70e-3} & 3 & 0.467\scriptsize{$\pm$3.33e-3} & 6 & 0.573\scriptsize{$\pm$3.45e-3} & 8 \\
        \hline
        Gauss & Argmax & 0.340\scriptsize{$\pm$2.51e-3} & 2 & \textbf{0.494}\scriptsize{$\pm$3.04e-3} & 1 & 0.618\scriptsize{$\pm$3.41e-3} & 5 \\
        Gauss & L1 & 0.326\scriptsize{$\pm$2.80e-3} & 4 & 0.491\scriptsize{$\pm$3.17e-3} & 2 & 0.631\scriptsize{$\pm$3.17e-3} & 2 \\
        Gauss & DP & 0.319\scriptsize{$\pm$2.52e-3} & 7 & 0.488\scriptsize{$\pm$2.62e-3} & 4 & \textbf{0.633}\scriptsize{$\pm$2.92e-3} & 1 \\
        \hline
        Prog-Bar & L1 & \textbf{0.341}\scriptsize{$\pm$2.38e-3} & 1 & 0.489\scriptsize{$\pm$2.34e-3} & 3 & 0.608\scriptsize{$\pm$2.71e-3} & 7 \\
        \hline
        Soft-Prog-Bar & L1 & 0.322\scriptsize{$\pm$4.04e-3} & 6 & 0.486\scriptsize{$\pm$4.32e-3} & 5 & 0.626\scriptsize{$\pm$4.62e-3} & 3 \\
        Soft-Prog-Bar & DP & 0.287\scriptsize{$\pm$2.11e-3} & 10 & 0.457\scriptsize{$\pm$3.13e-3} & 8 & 0.611\scriptsize{$\pm$4.09e-3} & 6 \\
        \hline
        Continuous & L1 & 0.290\scriptsize{$\pm$2.34e-3} & 9 & 0.463\scriptsize{$\pm$2.81e-3} & 7 & 0.620\scriptsize{$\pm$2.81e-3} & 4 \\
        \hline
        Bin-Num & L1 & 0.319\scriptsize{$\pm$2.63e-3} & 7 & 0.443\scriptsize{$\pm$3.10e-3} & 10 & 0.544\scriptsize{$\pm$3.73e-3} & 10 \\
        Bin-Num & DP & 0.324\scriptsize{$\pm$4.42e-3} & 5 & 0.452\scriptsize{$\pm$4.87e-3} & 9 & 0.559\scriptsize{$\pm$4.73e-3} & 9 \\
        \end{tabular}
        \caption[TODO caption]{
                \label{tab:pneumonicinfiltratesleftDeiT}pneumonic infiltrates, left; ViT-B-16; \#: relative rank; value\scriptsize{$\pm$ SD}
        }
    \end{center}
\end{table}

\pagebreak
\subsection{Pneumonic Infiltrates Right}

\begin{table}[h]
    \setlength{\tabcolsep}{2pt}
    \begin{center}
        \begin{tabular}{ l r | c c | c c | c c }
        Targ.-fn & Class.-fn & Unweight. $\kappa$ & \# & Lin. $\kappa$ & \# & Quad. $\kappa$ & \# \\
        \hline
        One-Hot & All & \textbf{0.401}\scriptsize{$\pm$1.63e-3} & 1 & 0.549\scriptsize{$\pm$2.20e-3} & 3 & 0.658\scriptsize{$\pm$2.65e-3} & 8 \\
        \hline
        Gauss & Argmax & 0.386\scriptsize{$\pm$2.24e-3} & 4 & 0.555\scriptsize{$\pm$3.24e-3} & 2 & 0.684\scriptsize{$\pm$3.69e-3} & 5 \\
        Gauss & L1 & 0.363\scriptsize{$\pm$7.32e-4} & 6 & 0.549\scriptsize{$\pm$1.73e-3} & 3 & \textbf{0.698}\scriptsize{$\pm$2.59e-3} & 1 \\
        Gauss & DP & 0.356\scriptsize{$\pm$4.17e-4} & 8 & 0.545\scriptsize{$\pm$1.29e-3} & 6 & \textbf{0.698}\scriptsize{$\pm$2.12e-3} & 1 \\
        \hline
        Prog-Bar & L1 & 0.391\scriptsize{$\pm$4.43e-3} & 2 & \textbf{0.556}\scriptsize{$\pm$3.89e-3} & 1 & 0.680\scriptsize{$\pm$3.51e-3} & 6 \\
        \hline
        Soft-Prog-Bar & L1 & 0.360\scriptsize{$\pm$3.37e-3} & 7 & 0.546\scriptsize{$\pm$3.74e-3} & 5 & 0.697\scriptsize{$\pm$4.05e-3} & 3 \\
        Soft-Prog-Bar & DP & 0.326\scriptsize{$\pm$4.94e-3} & 10 & 0.512\scriptsize{$\pm$4.93e-3} & 10 & 0.670\scriptsize{$\pm$4.97e-3} & 7 \\
        \hline
        Continuous & L1 & 0.334\scriptsize{$\pm$1.46e-3} & 9 & 0.527\scriptsize{$\pm$2.13e-3} & 7 & 0.692\scriptsize{$\pm$2.48e-3} & 4 \\
        \hline
        Bin-Num & L1 & 0.391\scriptsize{$\pm$1.36e-3} & 2 & 0.523\scriptsize{$\pm$1.27e-3} & 8 & 0.623\scriptsize{$\pm$1.88e-3} & 10 \\
        Bin-Num & DP & 0.380\scriptsize{$\pm$3.13e-3} & 5 & 0.521\scriptsize{$\pm$2.80e-3} & 9 & 0.635\scriptsize{$\pm$1.94e-3} & 9 \\
        \end{tabular}
        \caption[TODO caption]{
                \label{tab:pneumonicinfiltratesrightResNet50}pneumonic infiltrates, right; ResNet50; \#: relative rank; value\scriptsize{$\pm$ SD}
        }
    \end{center}
\end{table}

\begin{table}[h]
    \setlength{\tabcolsep}{2pt}
    \begin{center}
        \begin{tabular}{ l r | c c | c c | c c }
        Targ.-fn & Class.-fn & Unweight. $\kappa$ & \# & Lin. $\kappa$ & \# & Quad. $\kappa$ & \# \\
        \hline
        One-Hot & All & \textbf{0.366}\scriptsize{$\pm$4.41e-3} & 1 & 0.509\scriptsize{$\pm$4.47e-3} & 6 & 0.617\scriptsize{$\pm$4.62e-3} & 8 \\
        \hline
        Gauss & Argmax & 0.365\scriptsize{$\pm$1.89e-3} & 3 & \textbf{0.529}\scriptsize{$\pm$2.95e-3} & 1 & 0.656\scriptsize{$\pm$3.64e-3} & 4 \\
        Gauss & L1 & 0.342\scriptsize{$\pm$3.43e-3} & 6 & 0.522\scriptsize{$\pm$3.45e-3} & 3 & 0.668\scriptsize{$\pm$3.62e-3} & 2 \\
        Gauss & DP & 0.336\scriptsize{$\pm$3.42e-3} & 8 & 0.518\scriptsize{$\pm$3.29e-3} & 4 & \textbf{0.669}\scriptsize{$\pm$3.39e-3} & 1 \\
        \hline
        Prog-Bar & L1 & \textbf{0.366}\scriptsize{$\pm$2.90e-3} & 1 & 0.525\scriptsize{$\pm$2.96e-3} & 2 & 0.648\scriptsize{$\pm$3.05e-3} & 6 \\
        \hline
        Soft-Prog-Bar & L1 & 0.337\scriptsize{$\pm$3.60e-3} & 7 & 0.516\scriptsize{$\pm$4.01e-3} & 5 & 0.663\scriptsize{$\pm$4.27e-3} & 3 \\
        Soft-Prog-Bar & DP & 0.306\scriptsize{$\pm$2.43e-3} & 9 & 0.486\scriptsize{$\pm$3.01e-3} & 8 & 0.640\scriptsize{$\pm$3.63e-3} & 7 \\
        \hline
        Continuous & L1 & 0.306\scriptsize{$\pm$4.48e-3} & 9 & 0.495\scriptsize{$\pm$3.79e-3} & 7 & 0.656\scriptsize{$\pm$3.12e-3} & 4 \\
        \hline
        Bin-Num & L1 & 0.345\scriptsize{$\pm$3.11e-3} & 4 & 0.477\scriptsize{$\pm$2.85e-3} & 10 & 0.581\scriptsize{$\pm$2.91e-3} & 10 \\
        Bin-Num & DP & 0.343\scriptsize{$\pm$1.93e-3} & 5 & 0.479\scriptsize{$\pm$1.64e-3} & 9 & 0.591\scriptsize{$\pm$1.26e-3} & 9 \\
        \end{tabular}
        \caption[TODO caption]{
                \label{tab:pneumonicinfiltratesrightDeiT}pneumonic infiltrates, right; ViT-B-16; \#: relative rank; value\scriptsize{$\pm$ SD}
        }
    \end{center}
\end{table}

\pagebreak

\bibliographystyle{elsarticle-harv} 
\bibliography{regressionBib}

\end{document}